\documentclass[journal]{IEEEtran}
\usepackage[noend]{algpseudocode}
\usepackage{algorithmicx,algorithm}
\usepackage{bm}
\usepackage{graphicx}
\usepackage{subfigure}
\usepackage{amssymb}
\newcommand{\tabincell}[2]{\begin{tabular}{@{}#1@{}}#2\end{tabular}}
\usepackage[colorlinks, linkcolor=black, anchorcolor=black, citecolor=black]{hyperref}
\usepackage{multirow}

\ifCLASSINFOpdf
\else
\fi
\begin{document}
\title{Knowledge Distillation via Instance-level Sequence Learning}

\author{Haoran~Zhao, Xin~Sun, Junyu~Dong, Zihe~Dong and Qiong~Li
       
\thanks{This work was supported in part by National Natural Science Foundation of China under Project No.U1706218, 61971388 and L1824025. } 
%(Corresponding author: Xin Sun.) 

\thanks{H Zhao, X Sun, J Dong, C Chen, Z Dong and Q Li are with the Department of Computer Science and Technology, Ocean University of China, Qingdao, Shandong Province, 266100 China (e-mail:zhaohaoran@stu.ouc.edu.cn; sunxin@ouc.edu.cn; dongjunyun@ouc.edu.cn; dongzihe@stu.ouc.edu.cn; liqiong@stu.ouc.edu.cn)}

\thanks{Manuscript received April 19, 2005; revised August 26, 2015.}}


\markboth{Journal of \LaTeX\ Class Files,~Vol.~14, No.~8, August~2015}%
{Shell \MakeLowercase{\textit{et al.}}: Bare Demo of IEEEtran.cls for IEEE Journals}



\maketitle

\begin{abstract}
Recently, distillation approaches are suggested to extract general knowledge from a teacher network to guide a student network. Most of the existing methods transfer knowledge from the teacher network to the student via feeding the sequence of random mini-batches sampled uniformly from the data. Instead, we argue that the  compact  student  network should be guided gradually using samples ordered in a meaningful sequence. Thus, it can bridge the gap of feature representation between the teacher and student network step by step. In this work, we provide a curriculum learning knowledge distillation framework via instance-level sequence learning. It employs the student network of the early epoch as a snapshot to create a curriculum for the student network's next training phase. We carry out extensive experiments on CIFAR-10, CIFAR-100, SVHN and CINIC-10 datasets. Compared with several state-of-the-art methods, our framework achieves the best performance with fewer iterations.
\end{abstract}

\begin{IEEEkeywords}
Neural Networks Compression, Knowledge Distillation, Computer Vision, Deep Learning.
\end{IEEEkeywords}

% \fi
\IEEEpeerreviewmaketitle

\section{Introduction}
\IEEEPARstart
{D}{eep} neural networks (DNNs) have achieved superior performance on various computer vision tasks~\cite{Russakovsky2015ImageNet}\cite{6680765}\cite{6324460}\cite{7875137}. However, the promising results~\cite{7299291}\cite{6844850} come with the costs of the deeper and wider architectures, which require long inference time and high cost of computation. This drawback suppresses their implementation on devices with low memory and fast execution requirements such as mobile phones and embedded devices. Therefore, model compression techniques~\cite{Song2015Deep}~\cite{Hassibi1993Second}~\cite{Jaderberg2014Speeding}~\cite{Cun1989Optimal} have emerged to address such practical applications.

Knowledge distillation~\cite{Hinton2015Distilling}~\cite{Romero2015FitNets}~\cite{Zagoruyko2016Paying} provides an effective and promising solution for model compression. Generally speaking, the distilling technique utilizes a teacher-student framework to transfer knowledge from a complicated large teacher to a compact student. It is an effective approach to obtain a compact neural network with performance  close  to  the  complicated  teacher  network. It usually works by adding a distilling term to the original classification loss that encourages the student to mimic the teacher's behavior. For example, Hinton et al.~\cite{Hinton2015Distilling} propose to match the final predicted probabilities of the teacher and student network, as the distribution of the teacher network's soft targets contains more information than raw one-hot labels. Then Romero et al.~\cite{Romero2015FitNets} devise a hint-based training approach which matches intermediate layers of the student network to the corresponding layers of the teacher network. Zagoruyko et al.~\cite{Zagoruyko2016Paying} extend the hint-based idea to use attention as a mechanism for distilling knowledge from teacher network to student network.

\begin{figure}
  \centering
  \includegraphics[width=8.5cm]{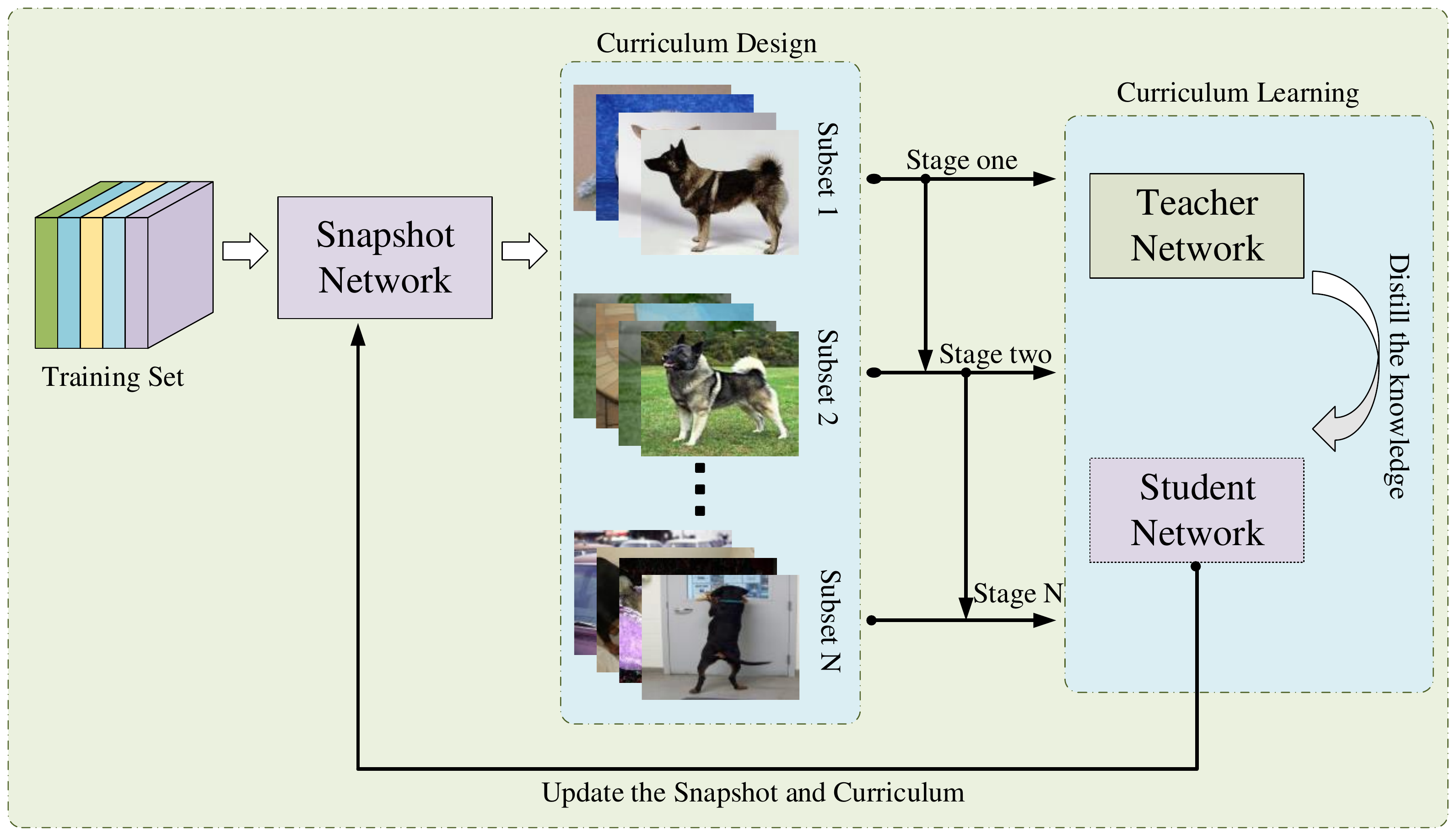}\\
  \caption{Pipeline of the instance-level sequence learning. The intermediate states of student network in the training process are employed to design curriculum for distilling knowledge. We adopt an easy-to-hard instance sequence learning to distill knowledge from the teacher to the student network through multistage training.}\label{fig:snapshot}
\end{figure}

However, we observe that the compact student network should be guided gradually using samples ordered in a meaningful sequence. Despite the teacher network conveys additional information beyond the conventional supervised learning target, the student network is harder to train than the teacher network due to the ability of feature representation. The conventional distillation methods usually train the networks by providing a sequence of random mini-batches sampled uniformly from the entire training data. Instead, we supervise the student network by generalizing easy samples before difficult ones. In other words, the teacher network guides the student network to construct an easy feature space using easy samples and then make it close to the teacher's complicated feature space by increasing the difficulty of training samples.

In this paper, we propose a new knowledge distillation framework called instance-level \textbf{S}equence \textbf{L}earning \textbf{K}nowledge \textbf{D}istillation (SLKD). It employs the early epochs as snapshot to create a curriculum for the student network's next training epochs. Figure \ref{fig:snapshot} shows the overall procedure. Curriculum learning, which is motivated by the idea of a curriculum in human learning, is employed in our distillation framework to supervise the student's training progress. In the first training phase, the student network is trained under the pre-trained teacher network by feeding the random mini-batches sampled uniformly from the entire training data. Then we use the early epochs of the student network as snapshot to make a curriculum by ranking the training data for the next training phase. Moreover, the easy samples from the resulting classifier are firstly feed to the teacher and student network in the knowledge distillation framework. Then difficult samples gradually join in the training. Therefore, the teacher network is no longer blindly supervising the students, but receiving feedback from the student network and advancing in regular order.

Our contributions in this paper are summarized as follows:

\begin{itemize}
\item We propose a novel teacher-student knowledge distillation framework with curriculum learning. The framework allows the student model to get generalization capability by continuously adding the training data with increasing complexity into the distillation process. 

\item We use snapshot of the student to rank sample complexity without involving extra training resource consumption. It makes the teacher efficiently supervise the student's training process by the student's feedback on performance.

\item We verify our proposed instance-level sequence learning knowledge distillation framework on CIFAR-10, CIFAR-100, SVHN and CINIC-10 datasets. Experiments show that our method can significantly improve the performance of student networks in knowledge distillation. 
\end{itemize}

The rest of this paper is organized as follows. Related work is reviewed in section \ref{section: sec2}, and we present the proposed instance-level sequence learning for knowledge distillation architecture in section \ref{section: sec3}. Experimental results are presented in section \ref{section: sec4}. Finally, section \ref{section: sec6} concludes this paper.

\section{Related Work}
\label{section: sec2}

Since this work focuses on training a compact network with high performance via knowledge distillation, we briefly review the related literature on model compression and acceleration and knowledge distillation. In addition, we also give a brief review of the recent advances in curriculum learning.

\textbf{DNNs compression and acceleration} are significant to the real-time applications which have drawn increasing attention in recent years. A straight way to boost the inference speed of DNNs is parameter pruning~\cite{Cun1989Optimal}~\cite{Hao2016Pruning}~\cite{Han2015Learning}\cite{8320372}, which removes redundant weights from the trained larger network to obtain a small one. Typically, it retains the accuracy of the larger model by setting the proper prune ratio. Another way is low-rank decomposition~\cite{Denil2013Predicting}~\cite{Kim2015Compression}, which tries to factorize parameter-heavy layers into multiple lightweight ones by using the matrix decomposition technique. However, parameter pruning and low-rank decomposition usually leads to large accuracy drops, thus fine-tuning is a must to alleviate those drops.    

\textbf{Knowledge distillation methods} usually utilize the teacher-student strategy to transfer knowledge from a pre-trained large teacher network to a compact student network. It can directly get a compact model that retains the accuracy of a large model for facilitating the deployment at the test time. Bucilua et al.~\cite{Bucilua:2006:MC:1150402.1150464} pioneer these series of methods in model compression. They attempt to compress the information from an ensemble of heterogeneous neural networks into a single neural network. Subsequently, Caruana et al.~\cite{Caruana2013Do} extend this method through forcing the wider and shallower student network to mimic the teacher network, using an L2 norm to penalize the difference between the student’s and teacher’s logits. More recently, Hinton et al.~\cite{Hinton2015Distilling} firstly propose the concept of knowledge distillation, which trains the student network by imitating the distribution of teacher network's soften output. The output from the teacher network, which divides the logits before softmax by introducing a hyper-parameter temperature, contains more information than one-hot targets.

Since then, researchers attempt to explore variants of knowledge distillation by using more supervised information from the teacher network. In~\cite{Romero2015FitNets}, Romero et al. introduce a new metric of intermediate features between the teacher and student networks and add a regressor to match different size of teacher's and student's outputs. Zagoruyko et al.~\cite{Zagoruyko2016Paying} propose to use the activation-based and gradient-based spatial attention maps from intermediate layers as the supervise information. Yim et al.~\cite{Yim2017A} propose to use the flow of solution procedure (FSP) that is generated by computing the Gram matrix of features between layers to transfer knowledge. And the students imitate the process of solving problems by the teachers in FSP method.

Different from the above methods, several researches adopts multiple teachers to supervise the student network's training. Shan et al.~\cite{Shan2017Learning} conduct distillation by combining the knowledge of intermediate representations from multiple teacher networks. Shen et al.~\cite{DBLP:journals/corr/abs-1811-02796} extend this idea by learning a compact student model which is capable of handling the super task from multiple teachers. 

Moreover, knowledge distillation can be used by combining with the conventional DNNs compression and acceleration approaches. In~\cite{Mishra2017Apprentice}, Mishra et al. propose a novel method to combine network quantization with knowledge distillation by jointly training a teacher network (full-precision) and a student network (low-precision) from scratch based on knowledge distillation. Besides, Zhou et al.~\cite{zhou2017Rocket} also propose a similar framework that the student network and the teacher network sharing lower layers and training simultaneously. Recently, researchers study knowledge distillation from another perspective rather than model compression. Born-again-network~\cite{DBLP:conf/icml/FurlanelloLTIA18} optimizes the same network in generations by training the students parameterized identically to their teachers. Yang et al.~\cite{DBLP:journals/corr/abs-1805-05551} optimize deep networks in many generations and a few networks with the same architecture are optimized one by one. A similar idea has been proposed to extract useful information from earlier epochs in the same generation~\cite{DBLP:journals/corr/abs-1812-00123}.

\begin{figure*}
  \centering
  \includegraphics[width=1\linewidth]{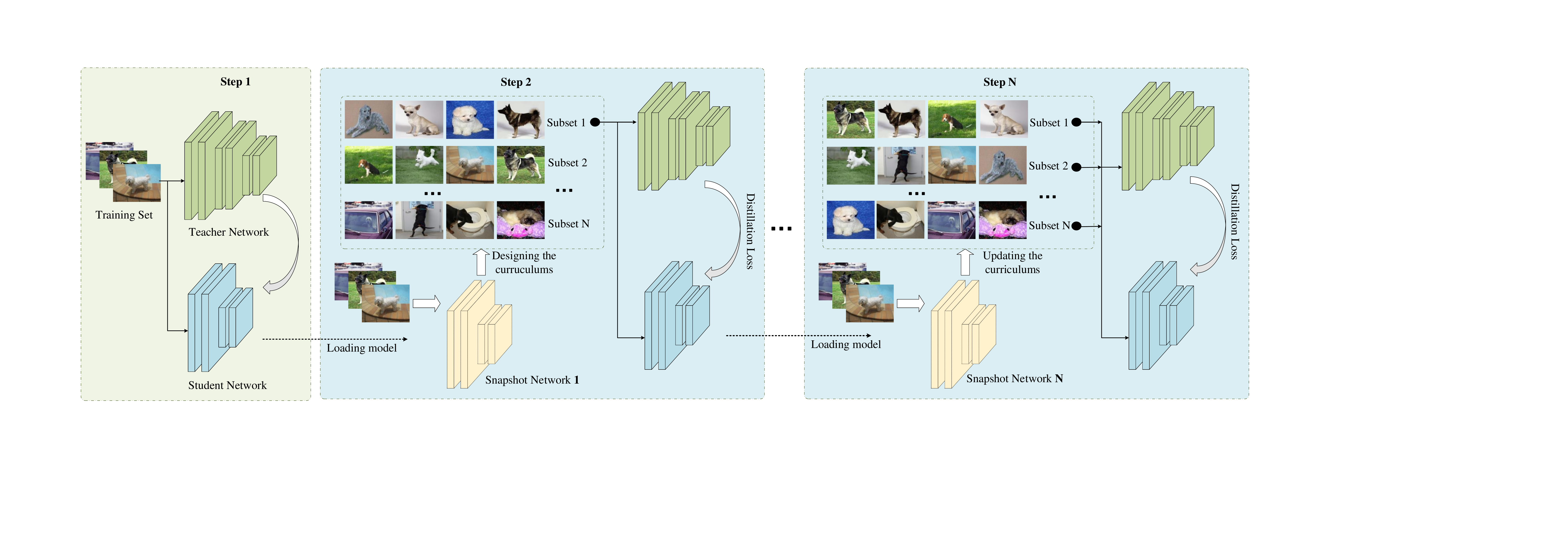}\\
  \caption{The overall framework of instance-level sequence learning for knowledge distillation. We get the first snapshot network from the student network through the conventional KD in Step 1. Then we design the easy-to-hard instance-level sequence curriculum via the snapshot network in Step 2. The Subset1, which is the easiest examples in the training set, is fed to distill knowledge from the teacher network to student network. In Step N, the whole training data is included for training the student network.}\label{fig:optimization}
\end{figure*}

As shown above, knowledge distillation typically transfers knowledge via feeding the sequence of random mini-batches sampled uniformly from the training data. In contrast, the student network in our proposed SLKD method is gradually guided using samples ordered in a meaningful sequence.

\textbf{Curriculum Learning} is a learning paradigm inspired by the cognitive process of human and animals, in which the model is trained gradually using samples ordered in a meaningful sequence. A
curriculum specifies a scheme under which training samples will be gradually learned. Bengio et al.~\cite{Bengio:2009:CL:1553374.1553380} pioneer these series of methods in curriculum learning. They propose that it is better to feed samples organized in a meaningful way than employ samples at random so that the low complex examples are presented first. Hacohen et al.~\cite{DBLP:conf/icml/HacohenW19} employ the curriculum learning on the training of deep networks and analyze the effect of curriculum learning. Jiang et al.~\cite{Jiang2018Mentornet} propose a novel method to learn data-driven curriculums for deep CNNs trained on corrupted labels.

\section{Method}
\label{section: sec3}

In this section, we will describe our main concept of knowledge distillation via instance-level sequence learning. Firstly, we introduce the motivation of our SLKD method. Then the following part describes the training objective and procedure.

\subsection{Motivation}

Knowledge distillation methods usually train the student to mimic the teacher's class probabilities or feature representation. Supervised by this softened knowledge, extra supervision information, such as the class correlation, is added to train the student network. However, the optimization of the student model is much harder due to the limited feature extraction capability. Moreover, the teacher network does not receive any feedback information from the student network in the whole distillation process.

\begin{table}[h]
\begin{center} 
\caption{ The accuracy and loss of student network trained with different lessons under the supervision of teacher network on CIFAR-100 dataset. 'Loss' represents the training loss.}
\label{table:lesson}
\begin{tabular}{|c|c|c|c|c|c|c|c|c|}
\hline
ResNet-20 &Lesson\#01&Lesson\#02&Lesson\#03 \\
\hline
\hline
Top-1(\%) & 53.84 & 52.42 & 50.91\\
\hline
Loss & 1.183 & 2.729 & 4.129 \\
\hline
\end{tabular}
\end{center}
\end{table}

From the perspective of curriculum learning~\cite{Bengio:2009:CL:1553374.1553380}, the teacher network should gradually distill knowledge to the student through feeding samples ordered in a meaningful sequence. We claim that the local minima can be promoted by the easy-to-hard learning process. To validate this assumption, we partition the CIFAR-100 into three subsets by the pre-trained teacher network's confidence score and use the classification accuracy and training loss to evaluate its difficulty. ResNet-110 is used as the teacher network and ResNet-20 as the student network. The teacher is trained using cross-entropy loss and the student network is trained by KD loss. Table \ref{table:lesson} shows the results. We can find that the less difficult lesson has lower training loss. It means that the complex lesson takes a difficult target for the student network to approach.

Based on this philosophy, we propose the instance-level SLKD method which employs curriculum learning to order the training samples by difficulty. Figure \ref{fig:optimization} shows the overview of the SLKD. We use the snapshot networks to represent the intermediate training states of the student network in our architecture. As can be seen from Figure \ref{fig:optimization}, we employ the snapshot network to design the lessons (subsets) from the whole training dataset. The curriculum is updated dynamically in the whole training process. The student network is firstly trained on the easiest lesson and we gradually increase the difficulty of lessons in the next training stage. Therefore our distilling is proceeded sequentially from the easy examples to the hard ones. In this manner, we hope to ease the learning process and help the student network reach an ideal local minimum.

\subsection{Formulation}

The idea of knowledge distillation is to train the student network not only via the true labels information but also by mimicking the teacher's class probabilities or feature representation (activations of hidden layers). In other words, the teacher network could provide valuable dark knowledge as extra supervisory information besides the ground-truth labels.

Let us denote $x$, $y$ as the input of the DNN and one-hot labels of our architecture. $\phi_{s}$ and $\phi_{t}$ represent the student network and teacher network with parameters $W_{s}$ and $W_{t}$ respectively. Given an input image $x$, the teacher network outputs the final predictions as $P_{T}$ which are obtained by applying the softmax function on the un-normalized log probability values $a_{{T}}$, i.e. $P_{T} = softmax(a_{{T}})$. Similarly, the same image is fed into the student network to obtain the predictions $P_{S} = softmax(a_{{S}})$. The standard cross-entropy is denoted as $\mathcal {H}$. In classical supervised learning, the mismatch between the output of the student network softmax and the ground-truth label $y$ is usually penalized using cross-entropy loss:

\begin{equation}
\mathcal{L}_{CE}(W_{s}) = \mathcal{H}(y_{true},P_{S})
\label{con:inventoryflow0}
\end{equation}

Hinton et al. ~\cite{Hinton2015Distilling} extend previous works ~\cite{Bucilua:2006:MC:1150402.1150464} by training a compact student network to mimic the output probability distribution of the teacher network. They name this informative and representative knowledge as dark knowledge and try to match the softened outputs of teacher and student via a KL-divergence loss:

\begin{equation}
\mathcal{L}_{KD}(W_{s}) = \tau^2{KL}(P_{T},P_{S})
\end{equation}

It contains the relative probabilities of incorrect classification results provided by teacher networks. When we perform knowledge distillation with a temperature parameter $\tau$, the student network will be trained to optimize the following loss function:

\begin{equation}
\mathcal{L}_{Student}(W_{s}) = (1-\lambda)\mathcal{L}_{CE} + \lambda\mathcal{L}_{KD}\label{con:inventoryflow}
\end{equation}

\noindent where $\lambda$ is a second hyper parameter controlling the trade-off between the two losses.

The teacher network is sometimes deeper and wider than the above approaches, but sometimes has the similar size as the student network~\cite{DBLP:conf/icml/FurlanelloLTIA18}\cite{DBLP:journals/corr/abs-1805-05551}\cite{DBLP:journals/corr/abs-1812-00123}. Snapshot Distillation~\cite{DBLP:journals/corr/abs-1812-00123} proposes to finish teacher-student optimization within one generation which acquires teacher information from the previous iterations of the same training process. Inspired by this, we propose to employ the snapshot of student from the previous epochs to design curriculum for efficient knowledge distillation.

\textbf{Instance-level sequence learning for knowledge distillation.} We partition the entire knowledge distillation process into $N$ mini-generations. During the first generation, the student network $\phi_{s}$ is trained from the true labels and the teacher network's supervisory information using the conventional distillation terms (using uniformly sampled mini-batches) by Eq.(\ref{con:inventoryflow}). Then, at each consecutive generation, the snapshot of student network from earlier epochs is employed to design curriculum for efficient distillation. In other words, we use the snapshot as a classifier and take its confidence score as the scoring function for each image.

Without loss of generality, we measure the complexity of sample $X=\{(x_{i},y_{i})^N_{i=1}\}$ by a scoring function defined as $f: X \rightarrow R$. The sample $(x_{i},y_{i})$ is more difficult than sample $(x_{j},y_{j})$ if $f(x_{i},y_{i}) > f(x_{j},y_{j})$. Here, we choose the classifier from snapshot as scoring function $f$. Thus, the training data is ranked by the difficulty of samples and sorted by the scoring function $f$. Then a sequence of subset $X_{1},...,X_{i} \subseteq X$ is divided by difficulty. Note that, it is important to keep the sample balanced with the same number of samples from each class as in the training set. Thus, we keep the same number of each class per subset to avoid bias. 

Then the optimization goal of SLKD in $i_{th}$ step is to minimize the following loss function:

\begin{eqnarray}
\mathcal{L}_{SLKD}(W_{s}, W_{t}) = \mathcal{H}(y_{true}, \phi_{s}(X_{i}, W_{s}) )
+\\\nonumber 
\lambda\mathcal{H}(\phi_{s}(X_{i}, W_{s}) , \phi_{t}(X_{i}, W_{t})) \label{con:inventoryflow1}
\end{eqnarray}

\noindent where the $\phi_{s}(X_{i}, W_{s})$ and $\phi_{t}(X_{i}, W_{t})$ refer to the corresponding outputs of student and teacher network, respectively. The parameter $W_{s}$ is optimized by learning to the instance-level easy-to-hard curriculums sequentially. 

\begin{algorithm}[t]
\caption{Training with Curriculum learning} 
\hspace*{-0.08cm} {\bf Input:} 
Teacher network $N_{T}$, Student network $N_{S}$, Snapshot model $M_{S}$ and training dataset, i.e., image data and label data $X=\{(x_{i},y_{i})^N_{i=1}\}$.\\
\hspace*{-0.08cm} {\bf Output:} 
parameters $W_{S}$ of student model.\\
\hspace*{-0.08cm} {\bf Initialize:} 
$W_{S}$, $W_{T}$ and training hyper-parameters.\\
\hspace*{-0.08cm} {\bf Stage 1:} 
Prepare the teacher network.
\begin{algorithmic}[1]
\State \bfseries Repeat: \mdseries
    \State compute $\mathcal{H}(y_{true},P_{T})$.
    \State update $ W_{T} $ by gradient back-propagation.
\State \bfseries Until: \mdseries $\mathcal{H}(y_{true},P_{T})$ converges.
\end{algorithmic}
\hspace*{-0.08cm} {\bf Stage 2:} 
Training the student via curriculum learning.
\begin{algorithmic}[1]
\State Loading the snapshot model of student and sort $X$ according to the classifier, i.e., $f$.
\State \bfseries Repeat: \mdseries
    \State feed subset $X_{1},...,X_{i} \subseteq X$ in order.
    \State compute $\mathcal{L}_{SLKD}(W_{s})$ by Eq. (\ref{con:inventoryflow1}).
    \State update $W_{s}$ by gradient back-propagation.
\State \bfseries Until: \mdseries $\mathcal{L}_{SLKD}(W_{s})$ converges.
\State \Return $ W_{s} $
\end{algorithmic}
\end{algorithm}

\subsection{Training procedure}

The training procedure is illustrated in Algorithm 1. In this paper, we partition the entire training procedure into $N$ phases. We update the teaching strategy (i.e., curriculum) at the beginning of every training phase to supervise the student network. 
Specifically, we firstly train the teacher network under the conventional supervised learning by Eq.(\ref{con:inventoryflow0}). 
To obtain an initial snapshot network, we distill the knowledge from the pre-trained teacher network to the student network by Eq.(\ref{con:inventoryflow}) through feeding a sequence of uniformly sampled mini-batches. Then, we load the student's model (checkpoints) from earlier epochs as snapshot which is used to design the curriculum for knowledge distillation. Thus, the whole training dataset is divided into several subsets according to the sample's difficulty. In each training phase, the teacher network supervises the training process of student network through different curriculums, which contains all the categories in the dataset. In this way, the student network is trained to generalize easy samples before the hard ones. The subset of training data is gradually added into the training process from easy to hard and eventually all the training data is included.

\begin{table*}
\begin{center}
\caption{Experiments on CIFAR-10 and CIFAR-100 datasets with different knowledge distillation methods. We conduct the experiments using the Resnet-110 as teacher network and the Resnet-20 as sutdent network. Recognition rates (\%) are computed as median of 5 runs with different seed.}
\label{table:1}
\begin{tabular}{|c|c|c|c|c|c|}
\hline
Method & Model & FLOPs(M) & Params(M) &  CIFAR-10(\%) & CIFAR-100(\%) \\
\hline
\hline
Student & Resnet-20 & 41.29 & 0.27 & 92.18 &68.33 \\
KD & Resnet-20 & 41.29 & 0.27 & 92.42 & 69.10 \\
FitNet & Resnet-20 & 41.29 & 0.27 & 92.55  & 69.48 \\
AT & Resnet-20 & 41.29 & 0.27 & 92.84  & 69.05 \\
Ours & Resnet-20 & 41.29 & 0.27 & \bfseries93.21\mdseries  & \bfseries70.02\mdseries \\
\hline
\hline
Teacher & Resnet-110 & 255.70 & 1.70 & 94.04 & 72.65\\
\hline
\end{tabular}
\end{center}
\end{table*}

At the beginning of next training phase, we load the new snapshot from earlier epochs and update the curriculum via the confidence scores of examples. The distillation process is the same as above. In this way, there is an interaction between the teacher network and student network in the knowledge distillation process via employing the snapshot of student. Moreover, this approach could improve the student network's generalization performance.

\section{Experiments}
\label{section: sec4}

In the following sections, we verify the effectiveness of our proposed  instance-level Sequence Learning Knowledge Distillation (SLKD) on several standard datasets, including CIFAR-10, CIFAR-100, SVHN and CINIC-10. For all experiments, a deep residual network~\cite{DBLP:conf/cvpr/HeZRS16} has been employed as the base architecture. The deep residual network has shortcut connections to make an ensemble structure. Furthermore, the shortcut connections allow training of very deep networks. Therefore, many researchers use the residual network for various tasks.

For all experiments, we compare our proposed method with several state-of-the-art knowledge distillation methods, including knowledge distillation (KD)~\cite{Hinton2015Distilling}, Attention Transfer Knowledge Distillation (ATKD)~\cite{Zagoruyko2016Paying} and Fitnets~\cite{DBLP:journals/corr/RomeroBKCGB14}.

\subsection{Experimental Setup}
\textbf{Network architecture.} For all experiments, we use the ResNet-110 as the teacher model, and the ResNet-20 is adopted as the student model. It stacks the basic residual blocks to achieve the state-of-the-art performance. To further investigate
the effectiveness of our method, we conduct extensive experiments on different network architecture families, such as MobileNet~\cite{DBLP:journals/corr/HowardZCKWWAA17} and ShuffleNetV2~\cite{Ma_2018_ECCV}. Table \ref{table:parameters} shows the number of parameters of the networks we adopt in our experiments for CIFAR-100 dataset.

\begin{table}[H]
\begin{center} 
\caption{Number of parameters on the CIFAR-100 dataset.}
\label{table:parameters}
\begin{tabular}{|c|c|c|c|c|c|c|c|c|}
\hline
ResNet-110 & ResNet-20 & MobileNet & ShuffleNetV2 \\
\hline
\hline
1.7M & 0.27M & 3.3M & 1.3M\\
\hline
\end{tabular}
\end{center}
\end{table}

\textbf{Implementation Details.} We firstly conduct our experiments on the public datasets CIFAR-10 which has $32\times32$ small RGB images. For all experiments, we use minibatches of size 128 for training. Moreover, we use horizontal flips and random crops for data augmentations before each minibatch. The learning rate start with 0.1 and is reduced by a factor of 0.1 on epoch 60, 120 and 160 respectively. For CIFAR dataset, we use stochastic gradient descent with momentum fixed at 0.9 for 200 epochs. However, we use Adam~\cite{DBLP:journals/corr/KingmaB14} with learning rate 0.01 initially and drop the learning rate by 0.2 at epoch 20, 40, 60 for SVHN dataset which is easy to learn. Furthermore, all networks have batch normalization.

For the baseline methods, we set the hyper-parameters as following. In the KD method, we set the temperature for softened softmax to 4 and $\lambda$ = 16~\cite{Hinton2015Distilling}. For the FitNet and AT methods, the value of $\lambda$ is set to $10^2$ and $10^3$~\cite{Zagoruyko2016Paying}. The mapping
function of AT adopted in our experiment is square sum, which performs the best in the experiments of ~\cite{Zagoruyko2016Paying}.

\subsection{CIFAR} 
In this section, we evaluate our method on the CIFAR dataset. The CIFAR-10 and CIFAR-100 datasets both contain 60k tiny RGB images at a spatial resolution of $32\times32$. The only difference is that both training and testing images are uniformly distributed over 10 or 100 classes, respectively. Note that, we use the $32\times32$ RGB images after random crops and horizontal flips for training and the original $32\times32$ RGB images are used for testing.
For optimization, we take SGD with a mini-batch size of 128. The learning rate starts from 
0.1 and is divided by 10 at 60th, 120th, 160th epochs and we train for 200 epochs.

Specifically, we firstly train the Resnet-110 as the teacher network which provides 94.04\% classification accuracy on CIFAR-10 dataset. Then, we use the pre-trained teacher network to supervise the training of student network for initial 40 epochs by Eq.(\ref{con:inventoryflow}). For CIFAR-10 dataset, we set $N=5$ in our experiment which partition the training data into 5 degree by the samples' difficulty. Note that, the five subsets of training data are provided by five different snapshots of the student network in the whole training process. In details, we load the first snapshot of student network from initial 40 epochs and use it as classifier to rank the whole training data into five subsets. In other words, this snapshot designs the first curriculum for next knowledge distillation training. Then, we fed the easiest subset to our knowledge distillation framework. The teacher network supervises the training of student network for next 30 epochs. In the same manner, the next snapshot is loaded to update the curriculum and the second subset is added into the training process for the next 30 epochs. Finally, the entire training data is included and we train the student for the last 100 epochs.

\begin{table*}
\begin{center} 
\caption{Recognition rates (\%) on SVHN and CINIC-10 datastes (The median accuracy over five runs is reported). Resnet-110 is as teacher network, Resnet-20 as student network. We compare different knowledge distillation methods such as traditional KD, attention transfer, and FitNet. The best result in each experiment is shown in bold.}
\label{table:cinic}
\begin{tabular}{|c|c|c|c|c|c|c|c|c|}
\hline
Dataset & Model(S/T) & Student &  KD  & AT & FitNet & Ours & Teacher\\
\hline
\hline
SVHN &\tabincell{c}{ResNet-20  \\ ResNet-110} & 94.98 & 95.43 & 95.81&95.97 &\bfseries96.25\mdseries &96.63  \\
\hline
\hline
CINIC-10 &\tabincell{c}{ResNet-20 \\ ResNet-110} & 82.43 & 82.58 & 82.84 &82.93 & \bfseries83.16\mdseries & 86.45 \\
\hline
\hline
\end{tabular}
\end{center}
\end{table*}

We compare the classification accuracy of CIFAR dataset and show the results in Table \ref{table:1}. From the results we can find that our approach achieves higher accuracy than the original student network and also gets notable improvement compared to the existing methods. Note that, our new architecture of instance-level sequence learning for knowledge distillation gets 93.21\% accuracy with 1.03\% improvement than the student network trained independently on CIFAR-10 dataset. In other words, the knowledge of the Resnet-110 teacher network, which contains 1.70$M$ parameters, is distilled into a smaller Resnet-20 student network with only 0.27$M$ parameters. This is a 6$X$ compression rate with only 0.8\% loss in accuracy.  Furthermore, our approach performs better than exiting knowledge distillation methods. For CIFAR-100 dataset, we change the parameter $N=3$ and partition the training data into $3$ subsets. Then, the snapshot of student is loaded at 41, 71, 141 epoch, respectively. And our method achieves 70.02\% classification accuracy on CIFAR-100 dataset and gets 1.69\% improvement compared with the student network trained individually. Moreover, our method outperforms the state-of-the-art knowledge distillation methods as shown in Table \ref{table:1}. 

\begin{figure}
  \centering
  \includegraphics[width=8.5cm]{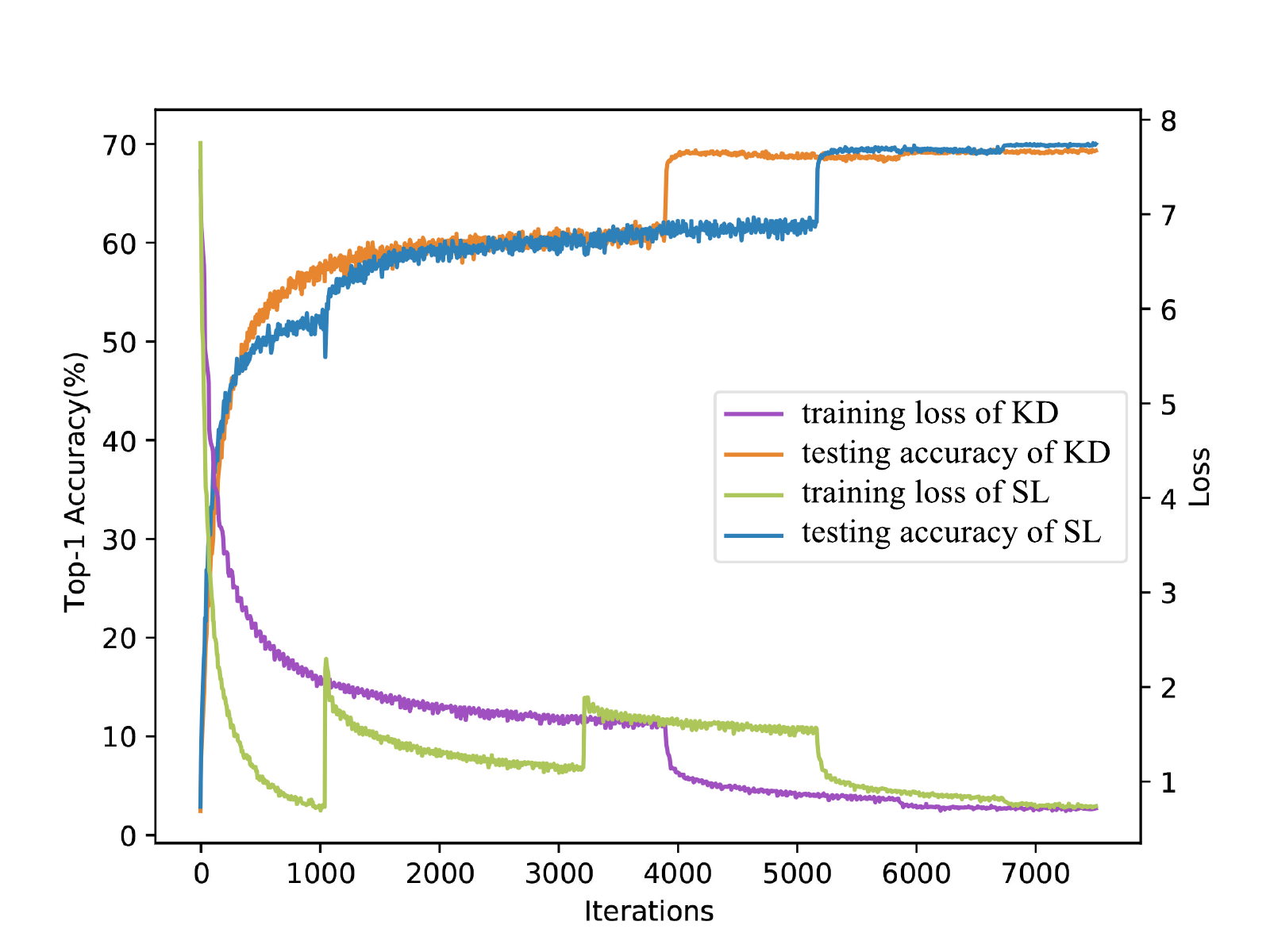}\\
  \caption{The training loss and testing accuracy of conventional KD and the proposed SLKD method on CIFAR-100 dataset. In this experiment, we adopt Resnet-110 as teacher network and Resnet-20 as student network. Both of them train on 7.5K iterations.}\label{fig:training loss}
\end{figure}

Figure \ref{fig:training loss} shows the training loss and testing accuracy of the conventional KD and our SLKD method. We train the network with the same iterations for comparison. As can be seen from the testing accuracy curves, our method outperforms the KD method on classification accuracy. Note that, the loss curves in our approach suddenly pulls up at 100 and 150 epochs because the new curriculum is added. The student network from our architecture finally reaches the saturation region and consequently results in a higher performance.

Actually, we conduct all the experiments on 300 epochs for CIFAR-10 and CIFAR-100 datasets. We can observe that our approach uses less iterations than the other comparative trials. Because we use the snapshot of student to design curriculum and feed the samples from easy to difficult ones in a proper order. That is to say, we only use a small subset of the entire training data in earlier epochs. We also set the same iterations (7.5K) for our experiments as Figure \ref{fig:training loss} shown. For designing curriculum, we attempt to investigate different curriculum designing strategy in section \ref{section: sec5}.

\subsection{SVHN and CINIC-10}

In this section, we verify the effectiveness of the proposed method through conducting complicated classification experiments on the larger SVHN and CINIC-10 datasets. The SVHN dataset is similar to MNIST with small $32\times32$ RGB cropped digits in 10 classes and it is obtained from house numbers in Google Street View images. SVHN has 73257 images for training, 26032 images in testing set and 531131 samples additional. The CINIC-10 dataset consists of images from both CIFAR dataset and ImageNet dataset, which is a middle option relative to CIFAR-10 and ImageNet. It contains 270,000 images at a spatial resolution of $32\times32$ via the addition of downsampled ImageNet images. We adopt the CINIC-10 dataset for rapid experimentation because its scale is closer to that of ImageNet and it is a noisy dataset. 

In this experiment, the backbone network of teacher is Resnet-110. For the training of teacher network on CINIC-10 dataset, we set initial learning rate to 0.1 and drop by 0.1 at 100, 150, 250 epochs and train for 300 epochs. We set the weight decay to 5e-4 and train the network using SGD with momentum. And the student network is set almost identical with the teacher except the initial learning rate is 0.01.

We compare the classification accuracy on SVHN, CINIC-10 dataset and show the results in Table \ref{table:cinic}. As can be seen from Table \ref{table:cinic}, our method improves about 1.27\%/0.73\% on SVHN/CINIC-10 compared with conventional cross-entropy supervised loss. We also compare our method with some state-of-the-art knowledge distillation methods. We can see that the student from our proposed method outperforms others. The results verify that our method is applicable to large classification.

\begin{table*}
\begin{center} 
\caption{ Student and teachers from different architecture families on CIFAR-100. Recognition rate (\%)(5 runs) is reported. The best result for each experiment is shown in bold. Brackets show the model size in number of parameters.}
\label{table:networks}
\begin{tabular}{|c|c|c|c|c|c|c|c|c|}
\hline
Student & Teacher & Student &  KD  & AT & FitNet & Ours & Teacher\\
\hline
\hline
MobileNet-0.25(0.2M) & ResNet-110(1.7M) & 78.54 & 79.63 & 79.95 & 81.29 & \bfseries81.45\mdseries & 83.84  \\
\hline
\hline
ShuffleNetV2-0.5(0.4M) & ResNet-110(1.7M) & 82.43 & 82.58 & 82.73 & 82.97 & \bfseries83.16\mdseries & 86.45 \\
\hline
\hline
\end{tabular}
\end{center}
\end{table*}

\begin{figure}[H]
  \centering
  \includegraphics[width=8.5cm]{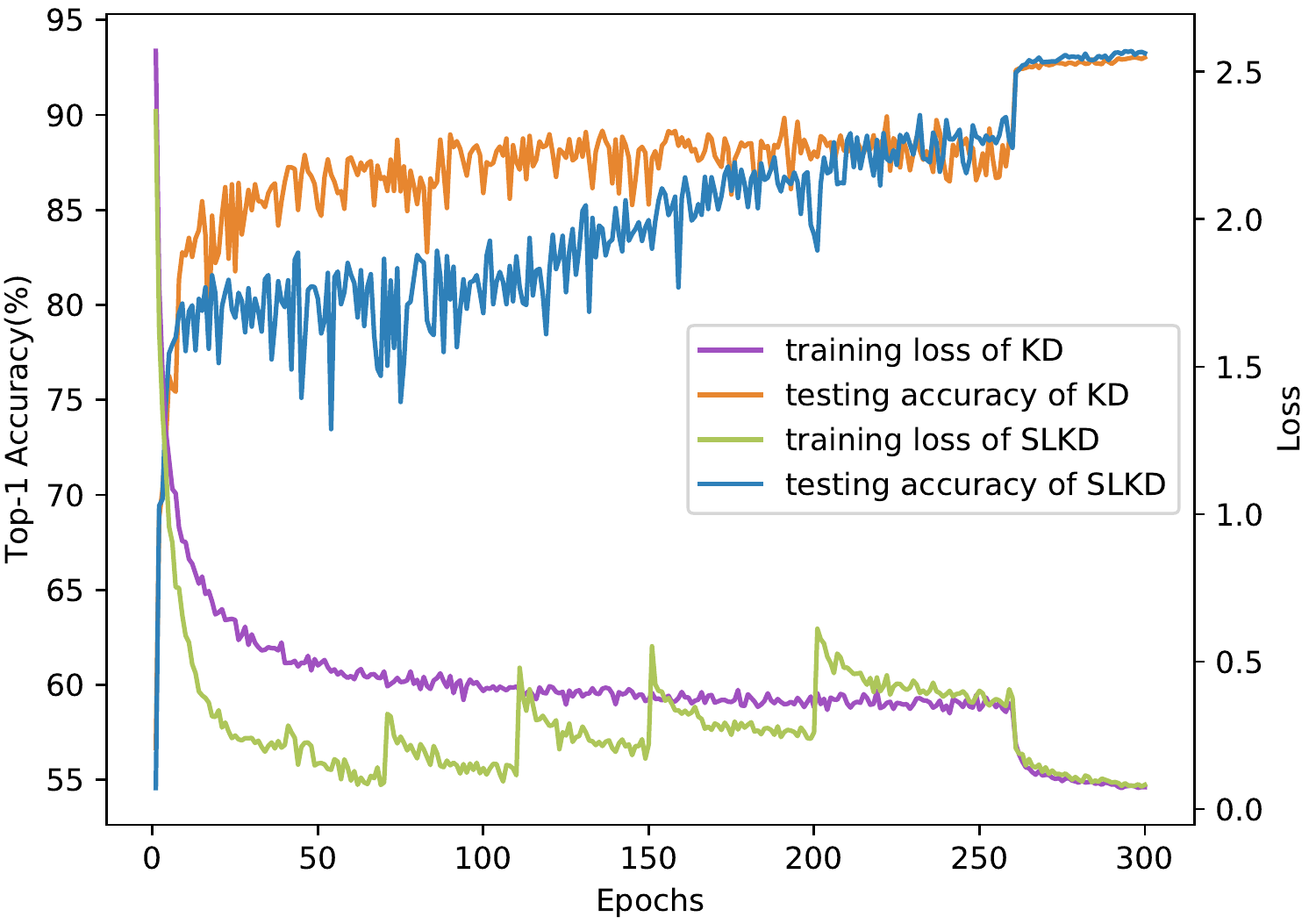}\\
  \caption{ The training loss and testing accuracy of conventional KD method and the proposed method on CIFAR-10 dataset. Note that, both methods train 300 epochs in this experiment.}\label{fig:epoch}
\end{figure}

\subsection{Ablation Studies}
\label{section: sec5}

With the easy-to-hard learning process inspired by curriculum learning, the student network can reach an ideal local minimum under the proposed instance-level sequence knowledge distillation method. The results in previous experiments have shown its effectiveness. In this section, we conduct extra experiments to investigate the impact of training time, different network architectures and snapshot networks on the proposed framework. 

\indent\textbf{Training with the same epochs.} Since training time plays a vital factor in either research or industrial, we aim to investigate the time consumption between the conventional KD method and the proposed method. As known to all, the same training epochs usually mean the same iterations because one training epoch contains the same iterations. However, the number of iterations in each epoch is different under our easy-to-hard sequence learning strategy. That is because the curriculum (subset) only contains $1/N$ examples of the whole training set in the early training stage. The number of iterations in each epoch increases when the new curriculum is added. In other words, our SLKD method outperforms other methods with fewer iterations when training with the same epochs. Figure \ref{fig:epoch} shows the training loss and testing accuracy of the conventional KD and the proposed method on 300 epochs. We can find that the proposed method outperforms the conventional KD method under the condition of using fewer total iterations. That means we use less training time than the conventional KD method.

\indent\textbf{Different student and teacher architectures.} To explore the impact of different architecture families, we conduct additional experiments on CIFAR-100 dataset. We adopt the ResNet-110 as the teacher network, MobileNet-0.25 and ShuffleNetV2-0.5 as the student network. The width multiplier is a tunable parameter to control the complexity of the ShuffleNetV2 and MobileNet. Table \ref{table:networks} shows the results under different architecture families. We can observe that the proposed approach outperforms the conventional training as well as the state-of-the-art knowledge distillation methods. The results verify that our approach can be applicable to different network architecture families.

\begin{table}
\begin{center} 
\caption{The recognition rates (\%) of the student network trained through different snapshot networks. "Snapshot-T" and "Snapshot-S" stand for the intermediate training states of teacher network and student network separately. The teacher network is ResNet-110, student network is ResNet-20.}
\label{table:ablation}
\begin{tabular}{|c|c|c|c|c|c|c|c|c|}
\hline
Snapshot & Network & CIFAR-10 & CIFAR-100 \\
\hline
\hline
Snapshot-T & \tabincell{c}{ResNet-20\\ ResNet-110}  & 92.03 & 68.11\\
\hline
Snapshot-S & \tabincell{c}{ResNet-20\\ ResNet-110}  & \bfseries93.21\mdseries & \bfseries70.02\mdseries \\
\hline
\end{tabular}
\end{center}
\end{table}

\indent\textbf{Comparison on different snapshots.} In this section , we evaluate the impacts of different curriculum designing strategy. In the proposed instance-level SLKD method, it employs the student network of the early epoch as the snapshot network to design the easy-to-hard curriculum for the student network's next training stage. Why we choose the intermediate training states of the student network as the snapshot networks? Typically, it is much easy to get the snapshots from the teacher network because it could produce tremendous checkpoints during the pre-trained process. However, our goal is to find the optimal learning sequence for distilling knowledge from the teacher network to the student network. If the curriculum is designed by the snapshot of teacher network, the student network may be misleaded due to the gap (in capacity) between the teacher and the student network. To validate the assumption, we compare our approach with another scheme, which using the pre-trained teacher network to provide the snapshot networks for curriculum designing. 

Specifically, we conduct extra experiment on the CIFAR-100 dataset. As comparison, we choose 60th, 120th, 160th epochs from the teacher network's training process as the snapshot networks. We load the snapshot networks from teacher network and student network respectively for designing the curriculum. The recognition results are shown in Table~\ref{table:ablation}. We can observe that the proposed method, which adopts the early epochs (checkpoints) of the student as the snapshot networks, outperforms the scheme using the teacher's intermediate states as snapshot networks.

\section{Conclusion}
\label{section: sec6}

In this paper, we propose a novel distillation framework named instance-level sequence learning knowledge distillation (SLKD) to boost the performance of student network. By employing the intermediate training states (i.e. checkpoints) of student network as snapshot networks, we design the easy-to-hard curriculums for distilling knowledge from teacher network to student. In such manner, we train the student network step-by-step to mimic the teacher's soften knowledge until finishing the whole curriculums. Moreover, the curriculums update automatically through the different snapshots in the whole training process. Experiments show that the proposed instance-level sequence learning strategy can significantly improve the performance of student networks in knowledge distillation.

\section*{Acknowledgment}
We thank supports of National Natural Science Foundation of China under Project No. U1706218, 61971388 and L1824025.

\bibliographystyle{IEEEtran}
\bibliography{IEEEexample}

\begin{thebibliography}{10}
\providecommand{\url}[1]{#1}
\csname url@samestyle\endcsname
\providecommand{\newblock}{\relax}
\providecommand{\bibinfo}[2]{#2}
\providecommand{\BIBentrySTDinterwordspacing}{\spaceskip=0pt\relax}
\providecommand{\BIBentryALTinterwordstretchfactor}{4}
\providecommand{\BIBentryALTinterwordspacing}{\spaceskip=\fontdimen2\font plus
\BIBentryALTinterwordstretchfactor\fontdimen3\font minus
  \fontdimen4\font\relax}
\providecommand{\BIBforeignlanguage}[2]{{%
\expandafter\ifx\csname l@#1\endcsname\relax
\typeout{** WARNING: IEEEtran.bst: No hyphenation pattern has been}%
\typeout{** loaded for the language `#1'. Using the pattern for}%
\typeout{** the default language instead.}%
\else
\language=\csname l@#1\endcsname
\fi
#2}}
\providecommand{\BIBdecl}{\relax}
\BIBdecl

\bibitem{Russakovsky2015ImageNet}
O.~Russakovsky, J.~Deng, H.~Su, J.~Krause, S.~Satheesh, S.~Ma, Z.~Huang,
  A.~Karpathy, A.~Khosla, and M.~Bernstein, ``Imagenet large scale visual
  recognition challenge,'' \emph{International Journal of Computer Vision},
  vol. 115, no.~3, pp. 211--252, 2015.

\bibitem{6680765}
G.~{Zhu}, Q.~{Wang}, and Y.~{Yuan}, ``Natas: Neural activity trace aware
  saliency,'' \emph{IEEE Transactions on Cybernetics}, pp. 1014--1024, 2014.

\bibitem{6324460}
Q.~{Wang}, Y.~{Yuan}, P.~{Yan}, and X.~{Li}, ``Saliency detection by
  multiple-instance learning,'' \emph{IEEE Transactions on Cybernetics},
  vol.~43, pp. 660--672, 2013.

\bibitem{7875137}
L.~{Wang}, H.~{Lu}, and M.~{Yang}, ``Constrained superpixel tracking,''
  \emph{IEEE Transactions on Cybernetics}, vol.~48, no.~3, pp. 1030--1041,
  2018.

\bibitem{7299291}
Y.~{Yuan}, J.~{Lin}, and Q.~{Wang}, ``Hyperspectral image classification via
  multitask joint sparse representation and stepwise mrf optimization,''
  \emph{IEEE Transactions on Cybernetics}, vol.~46, pp. 2966--2977, 2016.

\bibitem{6844850}
Y.~{Yuan}, J.~{Fang}, and Q.~{Wang}, ``Online anomaly detection in crowd scenes
  via structure analysis,'' \emph{IEEE Transactions on Cybernetics}, pp.
  548--561, 2015.

\bibitem{Song2015Deep}
H.~Song, H.~Mao, and W.~J. Dally, ``Deep compression: Compressing deep neural
  networks with pruning, trained quantization and huffman coding,''
  \emph{Fiber}, vol.~56, no.~4, pp. 3--7, 2015.

\bibitem{Hassibi1993Second}
B.~Hassibi and D.~G. Stork, ``Second order derivatives for network pruning:
  Optimal brain surgeon,'' \emph{Advances in Neural Information Processing
  Systems}, vol.~5, pp. 164--171, 1993.

\bibitem{Jaderberg2014Speeding}
M.~Jaderberg, A.~Vedaldi, and A.~Zisserman, ``Speeding up convolutional neural
  networks with low rank expansions,'' \emph{Computer Science}, vol.~4, no.~4,
  p. XIII, 2014.

\bibitem{Cun1989Optimal}
Y.~L. Cun, J.~S. Denker, and S.~A. Solla, ``Optimal brain damage,'' in
  \emph{International Conference on Neural Information Processing Systems},
  1989.

\bibitem{Hinton2015Distilling}
G.~Hinton, O.~Vinyals, and J.~Dean, ``Distilling the knowledge in a neural
  network,'' \emph{Computer Science}, vol.~14, no.~7, pp. 38--39, 2015.

\bibitem{Romero2015FitNets}
A.~Romero, N.~Ballas, S.~E. Kahou, A.~Chassang, and Y.~Bengio, ``Fitnets: Hints
  for thin deep nets,'' \emph{Computer Science}, 2015.

\bibitem{Zagoruyko2016Paying}
S.~Zagoruyko and N.~Komodakis, ``Paying more attention to attention: Improving
  the performance of convolutional neural networks via attention transfer,''
  \emph{arXiv preprint arXiv:1612.03928}, 2016.

\bibitem{Hao2016Pruning}
L.~Hao, A.~Kadav, I.~Durdanovic, H.~Samet, and H.~P. Graf, ``Pruning filters
  for efficient convnets,'' \emph{arXiv preprint arXiv:1608.08710}, 2016.

\bibitem{Han2015Learning}
S.~Han, J.~Pool, J.~Tran, and W.~Dally, ``Learning both weights and connections
  for efficient neural network,'' in \emph{Advances in neural information
  processing systems}, 2015, pp. 1135--1143.

\bibitem{8320372}
S.~{Lin}, J.~{Zeng}, and X.~{Zhang}, ``Constructive neural network learning,''
  \emph{IEEE Transactions on Cybernetics}, vol.~49, no.~1, pp. 221--232, Jan
  2019.

\bibitem{Denil2013Predicting}
M.~Denil, B.~Shakibi, L.~Dinh, M.~Ranzato, and N.~D. Freitas, ``Predicting
  parameters in deep learning,'' in \emph{International Conference on Neural
  Information Processing Systems}, 2013.

\bibitem{Kim2015Compression}
Y.~D. Kim, E.~Park, S.~Yoo, T.~Choi, Y.~Lu, and D.~Shin, ``Compression of deep
  convolutional neural networks for fast and low power mobile applications,''
  \emph{Computer Science}, vol.~71, no.~2, pp. 576--584, 2015.

\bibitem{Bucilua:2006:MC:1150402.1150464}
C.~Bucilu\v{a}, R.~Caruana, and A.~Niculescu-Mizil, ``Model compression,'' in
  \emph{Proceedings of the 12th ACM SIGKDD International Conference on
  Knowledge Discovery and Data Mining}, ser. KDD '06.\hskip 1em plus 0.5em
  minus 0.4em\relax New York, NY, USA: ACM, 2006, pp. 535--541.

\bibitem{Caruana2013Do}
J.~B. Lei and R.~Caruana, ``Do deep nets really need to be deep?''
  \emph{Advances in Neural Information Processing Systems}, pp. 2654--2662,
  2013.

\bibitem{Yim2017A}
J.~Yim, D.~Joo, J.~Bae, and J.~Kim, ``A gift from knowledge distillation: Fast
  optimization, network minimization and transfer learning,'' in \emph{IEEE
  Conference on Computer Vision \& Pattern Recognition}, 2017.

\bibitem{Shan2017Learning}
Y.~Shan, X.~Chang, X.~Chao, and D.~Tao, ``Learning from multiple teacher
  networks,'' in \emph{Acm Sigkdd International Conference on Knowledge
  Discovery \& Data Mining}, 2017.

\bibitem{DBLP:journals/corr/abs-1811-02796}
C.~Shen, X.~Wang, J.~Song, L.~Sun, and M.~Song, ``Amalgamating knowledge
  towards comprehensive classification,'' \emph{CoRR}, vol. abs/1811.02796,
  2018.

\bibitem{Mishra2017Apprentice}
A.~Mishra and D.~Marr, ``Apprentice: Using knowledge distillation techniques to
  improve low-precision network accuracy,'' \emph{arXiv preprint
  arXiv:1711.05852}, 2017.

\bibitem{zhou2017Rocket}
G.~Zhou, Y.~Fan, R.~Cui, W.~Bian, X.~Zhu, and G.~Kun, ``Rocket launching: A
  unified and effecient framework for training well-behaved light net,''
  \emph{arXiv preprint arXiv:1708.04106}, 2017.

\bibitem{DBLP:conf/icml/FurlanelloLTIA18}
T.~Furlanello, Z.~C. Lipton, M.~Tschannen, L.~Itti, and A.~Anandkumar,
  ``Born-again neural networks,'' in \emph{Proceedings of the 35th
  International Conference on Machine Learning, {ICML} 2018,
  Stockholmsm{\"{a}}ssan, Stockholm, Sweden, July 10-15, 2018}, 2018, pp.
  1602--1611.

\bibitem{DBLP:journals/corr/abs-1805-05551}
C.~Yang, L.~Xie, S.~Qiao, and A.~L. Yuille, ``Knowledge distillation in
  generations: More tolerant teachers educate better students,'' \emph{CoRR},
  vol. abs/1805.05551, 2018.

\bibitem{DBLP:journals/corr/abs-1812-00123}
C.~Yang, L.~Xie, C.~Su, and A.~L. Yuille, ``Snapshot distillation:
  Teacher-student optimization in one generation,'' \emph{CoRR}, vol.
  abs/1812.00123, 2018.

\bibitem{Bengio:2009:CL:1553374.1553380}
Y.~Bengio, J.~Louradour, R.~Collobert, and J.~Weston, ``Curriculum learning,''
  in \emph{Proceedings of the 26th Annual International Conference on Machine
  Learning}, 2009, pp. 41--48.

\bibitem{DBLP:conf/icml/HacohenW19}
G.~Hacohen and D.~Weinshall, ``On the power of curriculum learning in training
  deep networks,'' in \emph{Proceedings of the 36th International Conference on
  Machine Learning, {ICML} 2019, 9-15 June 2019, Long Beach, California,
  {USA}}, 2019, pp. 2535--2544.

\bibitem{Jiang2018Mentornet}
L.~Jiang, Z.~Zhou, T.~Leung, L.~J. Li, and F.~F. Li, ``Mentornet: Learning
  data-driven curriculum for very deep neural networks on corrupted labels,''
  in \emph{ICML 2018}, 2018.

\bibitem{DBLP:conf/cvpr/HeZRS16}
K.~He, X.~Zhang, S.~Ren, and J.~Sun, ``Deep residual learning for image
  recognition,'' in \emph{2016 {IEEE} Conference on Computer Vision and Pattern
  Recognition, {CVPR} 2016, Las Vegas, NV, USA, June 27-30, 2016}, 2016, pp.
  770--778.

\bibitem{DBLP:journals/corr/RomeroBKCGB14}
A.~Romero, N.~Ballas, S.~E. Kahou, A.~Chassang, C.~Gatta, and Y.~Bengio,
  ``Fitnets: Hints for thin deep nets,'' in \emph{3rd International Conference
  on Learning Representations, {ICLR} 2015, San Diego, CA, USA, May 7-9, 2015,
  Conference Track Proceedings}, 2015.

\bibitem{DBLP:journals/corr/HowardZCKWWAA17}
A.~G. Howard, M.~Zhu, B.~Chen, D.~Kalenichenko, W.~Wang, T.~Weyand,
  M.~Andreetto, and H.~Adam, ``Mobilenets: Efficient convolutional neural
  networks for mobile vision applications,'' \emph{CoRR}, vol. abs/1704.04861,
  2017.

\bibitem{Ma_2018_ECCV}
N.~Ma, X.~Zhang, H.-T. Zheng, and J.~Sun, ``Shufflenet v2: Practical guidelines
  for efficient cnn architecture design,'' in \emph{The European Conference on
  Computer Vision (ECCV)}, September 2018.

\bibitem{DBLP:journals/corr/KingmaB14}
D.~P. Kingma and J.~Ba, ``Adam: {A} method for stochastic optimization,''
  \emph{CoRR}, vol. abs/1412.6980, 2014.

\end{thebibliography}

\end{document}